\documentclass[11pt]{article}

\usepackage{colacl}
\title{Statistical Models for Unsupervised Prepositional Phrase Attachment\\}
\author{Adwait Ratnaparkhi\\
Dept.~of Computer and Information Science\\
University of Pennsylvania\\
200 South 33rd Street\\
Philadelphia, PA 19104-6389\\{\tt adwait@unagi.cis.upenn.edu}}

\begin{document}
\maketitle
\abstract{
We  present  several  unsupervised  statistical models for  the  prepositional  phrase
attachment task that approach the accuracy of the best supervised methods for this task. Our
unsupervised approach uses a heuristic based on attachment proximity and trains from raw text
that is annotated with only part-of-speech tags and morphological base forms, 
as opposed to attachment information. 
It is therefore less resource-intensive and more portable than previous corpus-based algorithm
proposed for this task. 
We present results for prepositional phrase attachment in both English and Spanish.
}

\bibliographystyle{acl}

\newcommand{\unknown}{?}
\newcommand{\tp}[1]{{\tt #1}}
\section{Introduction}

Prepositional phrase attachment is
the task of deciding, for a given preposition
in a sentence, the attachment site that
corresponds to the interpretation of the sentence.
For example, the task in the following examples
is to decide whether the preposition {\em with} modifies
the preceding noun phrase (with head word {\em shirt}) or the
preceding verb phrase (with head word {\em bought} or {\em washed}).
\begin{enumerate}
\item I bought the shirt with pockets.
\label{npattach}
\item I washed the shirt with soap.
\label{vpattach}
\end{enumerate}
In sentence \ref{npattach}, {\em with} modifies the noun {\em shirt}, 
since {\em with pockets} describes the {\em shirt}.
However in sentence \ref{vpattach}, {\em with} modifies
the verb {\em washed} since {\em with soap} describes
how the shirt is {\em washed}. 
While this form of attachment ambiguity is usually
easy for people to resolve, 
a computer requires detailed knowledge about
words (e.g., {\em washed} vs. {\em bought}) 
in order to successfully resolve such 
ambiguities and predict the correct
interpretation.


\section{Previous Work}

Most of the previous successful approaches to this problem have
been statistical or corpus-based, and they
consider only prepositions whose
attachment is ambiguous between a preceding noun phrase
and verb phrase.
Previous work has framed the problem as a classification
task, in which the goal is to predict $N$ or $V$, 
corresponding to noun or verb attachment, given
the head verb $v$, the head noun $n$, the preposition $p$,
and optionally, the object of the preposition $n2$.
For example, the $(v,n,p,n2)$ tuples corresponding 
to the example sentences are
\begin{enumerate}
\item bought shirt with pockets
\item washed shirt with soap
\end{enumerate}
The correct classifications of tuples 1 and 2 are
$N$ and $V$, respectively.

\cite{hindlerooth:cl} describes a partially supervised approach in which the
 FIDDITCH partial parser was used to extract
$(v, n, p)$ tuples from raw text, where $p$ is a preposition
whose attachment is ambiguous between the head verb $v$ and the 
head noun $n$. 
The extracted tuples are then used to construct a classifier, which resolves
 unseen ambiguities at around 80\% accuracy.
Later work, such as 
\cite{maxent:ppattach,brillresnik:pp,collins:pp,merlo:pp,zavrel:pp,franz:pp},
trains and tests on  quintuples of the form $(v,n,p,n2,a)$ extracted
from the Penn treebank\cite{treebank}, 
and has gradually improved on this accuracy 
with other kinds of statistical learning methods,
yielding up to 84.5\% accuracy\cite{collins:pp}.
Recently, \cite{stetina:pp} have reported 88\% accuracy 
by using a corpus-based model in conjunction
with a semantic dictionary. 

While previous corpus-based methods are highly accurate for this task,
they are difficult to port to other languages because they require resources 
that are expensive to construct or simply nonexistent in other languages.
We present an unsupervised algorithm for prepositional
phrase attachment in English that  requires
only an part-of-speech tagger and a morphology database, 
and is therefore less resource-intensive and more portable than previous approaches, which have all
required either treebanks or partial parsers.

\section{Unsupervised Prepositional Phrase Attachment}

The exact task of our algorithm will be to 
construct a classifier $cl$
which maps an instance of an ambiguous prepositional phrase 
$(v,n,p,n2)$ to either $N$ or $V$, corresponding to noun attachment
or verb attachment, respectively.
In the full natural language parsing task, there are more 
than just two potential attachment sites, 
but we limit our task to choosing between a verb $v$
and a noun $n$ so that we may compare with 
previous supervised attempts on this problem.
While we will be given the candidate attachment
sites during testing, the training procedure
assumes no a priori information  about potential
attachment sites.

\subsection{Generating Training Data From Raw Text}

\newcommand\nullword{?} 

\begin{table*}
\begin{center}
\begin{tabular}{|c|p{5.0in}|} \hline
Tool&Output\\ \hline \hline
Raw Text&The professional conduct of lawyers in other jurisdictions is guided by American Bar Association rules or by state bar ethics codes , none of which permit non-lawyers to be partners in law firms .\\
$\Downarrow$&\\
POS Tagger&The/DT professional/JJ conduct/NN of/IN lawyers/NNS in/IN other/JJ jurisdictions/NNS is/VBZ guided/VBN by/IN American/NNP Bar/NNP Association/NNP rules/NNS or/CC by/IN state/NN bar/NN ethics/NNS codes/NNS ,/, none/NN of/IN which/WDT permit/VBP non-lawyers/NNS to/TO be/VB partners/NNS in/IN law/NN firms/NNS ./. \\
$\Downarrow$&\\
Chunker& conduct/NN of/IN lawyers/NNS in/IN jurisdictions/NNS is/VBZ guided/VBN by/IN rules/NNS or/CC by/IN codes/NNS ,/, none/NN of/IN which/WDT permit/VBP non-lawyers/NNS to/TO be/VB partners/NNS in/IN firms/NNS ./. \\
$\Downarrow$&\\
Extraction Heuristic&($n=$lawyers, $p=$in, $n2=$jurisdictions)\\
&($v=$guided, $p=$by, $n2=$rules)\\
$\Downarrow$&\\
Morphology&($n=$lawyer, $p=$in, $n2=$jurisdiction)\\
&($v=$guide, $p=$by, $n2=$rule)\\ \hline
\end{tabular}
\caption{How to obtain training data from raw text}
\label{table:order}
\end{center}
\end{table*}

We generate training data from raw text
by using a part-of-speech tagger, a simple chunker, 
an extraction heuristic, and a morphology database.
The order in which these tools are applied to raw text is shown 
in Table~\ref{table:order}.
The tagger from \cite{maxent:tagger}  first  annotates sentences of raw text with a
sequence of part-of-speech tags.
The chunker, implemented with two small regular expressions, then replaces simple noun phrases and
quantifier phrases with their head words.
The extraction heuristic then finds 
head word tuples and their likely attachments from the tagged and chunked text. 
The heuristic relies on the observed fact that in English and in languages with similar word order,
the attachment site of a preposition is usually located only a few words to the
left of the preposition.
Finally, numbers are replaced by a single token, the text is converted
to lower case, and the morphology database is used to find the 
base forms of the verbs and nouns.

The extracted head word tuples differ from the training data 
used in previous supervised attempts in an important way.
In the supervised case, both of the potential sites, namely
the verb $v$ and the noun $n$ are known before the attachment
is resolved.
In the unsupervised case discussed here, 
the extraction heuristic only finds what it thinks are {\em unambiguous} cases
of prepositional phrase attachment. 
Therefore, there is only one possible attachment site for the preposition,
and either the verb $v$ or the noun $n$ does not exist,
in the case of noun-attached preposition or a verb-attached preposition,
respectively.
This extraction heuristic loosely resembles a
step in the bootstrapping procedure used to get
training data for the classifier of \cite{hindlerooth:cl}.
In that step, unambiguous attachments from the FIDDITCH parser's output are initially used to resolve
some of the ambiguous attachments, and the resolved cases
are iteratively used to disambiguate the remaining unresolved cases.
Our procedure differs critically from \cite{hindlerooth:cl} in that
we do not iterate, 
we extract unambiguous attachments from unparsed
input sentences,
and we totally ignore the ambiguous cases.
It is the hypothesis of this approach that the information
in {\em just the unambiguous} attachment events can resolve the {\em ambiguous} attachment
events of the test data.

\subsubsection{Heuristic Extraction of Unambiguous Cases}

Given a tagged and chunked sentence, the extraction heuristic 
returns  head word tuples of the form
$(v, p, n2)$ or $(n, p, n2)$,  where  $v$  is  the  verb,  $n$  is  the
noun, $p$ is the preposition, $n2$ is the object of the preposition.
The main idea of the extraction heuristic is that
an attachment site of a preposition is usually 
within a few words to the left of the preposition.
We extract :
\begin{description}
\item[$(v,p,n2)$] if
\begin{itemize}
\item $p$ is a preposition ($p \neq of$)
\item $v$ is the first verb that occurs within $K$ words to the left of $p$
\item $v$ is not a form of the verb {\em to be}
\item No noun occurs between $v$ and $p$
\item $n2$ is the first noun that occurs within $K$ words to the right of $p$
\item No verb occurs between $p$ and $n2$
\end{itemize}
\item[$(n,p,n2)$] if
\begin{itemize}
\item $p$ is a preposition ($p \neq of$)
\item $n$ is the first noun that occurs within $K$ words to the left of $p$
\item No verb occurs within $K$ words to the left of $p$
\item $n2$ is the first noun that occurs within $K$ words to the right of $p$
\item No verb occurs between $p$ and $n2$
\end{itemize}
\end{description}
Table~\ref{table:order} also shows the result of the
applying the extraction heuristic to a sample sentence.

The heuristic ignores cases where $p=of$, since such cases
are rarely ambiguous, and we opt to model them 
deterministically as noun attachments. We will report
accuracies (in Section~\ref{evalsec}) 
on both cases where $p=of$ and where $p \neq of$.
Also, the heuristic excludes examples with the verb {\em to be} 
from the training set (but not the test set) since we found them
to be unreliable sources of evidence.

\subsection{Accuracy of Extraction Heuristic}

Applying the extraction heuristic to 970K 
unannotated sentences
from the 1988 Wall St. Journal\footnote{This
data is available from the Linguistic Data
Consortium, {\tt http://www.ldc.upenn.edu}}
data yields
approximately 910K unique head word
tuples of the form $(v,p,n2)$ or $(n,p,n2)$.
The extraction heuristic is far from perfect;
when applied to and compared with the annotated Wall St. Journal data of the
Penn treebank, only 69\% of the extracted
head word tuples represent correct attachments.\footnote{This
accuracy  also excludes cases where $p=of$.}	
The extracted tuples are meant to be 
a noisy but {\em abundant} substitute
for the information that one might
get from a treebank.
Tables \ref{vtuples} and \ref{ntuples} list the
most frequent extracted head word tuples for
unambiguous verb and noun attachments, respectively.
Many of the frequent noun-attached $(n,p,n2)$ tuples,
such as  {\em num to num},\footnote{Recall the
{\em num} is the token for quantifier phrases identified by the chunker,  like
{\em 5 million}, or {\em 6 \%}.}
are incorrect.
The prepositional phrase
{\em to num} is usually attached
to a verb such as {\em rise} or {\em fall}
in the Wall St. Journal domain, e.g., 
{\em Profits rose 46 \% to 52 million.}

\begin{table}
\begin{center}
\begin{tabular}{|c|c|c|c|} \hline
Frequency&Verb&Prep&Noun2\\ \hline \hline
8110&close&at&num\\ \hline
1926&reach&for&comment\\ \hline
1539&rise&to&num\\ \hline
1438&compare&with&num\\ \hline
1072&fall&to&num\\ \hline
970&account&for&num\\ \hline
887&value&at&million\\ \hline
839&say&in&interview\\ \hline
680&compare&with&million\\ \hline
673&price&at&num\\ \hline
\end{tabular}
\caption{Most frequent $(v,p,n2)$  tuples}
\label{vtuples}
\end{center}
\end{table}

\begin{table}
\begin{center}
\begin{tabular}{|c|c|c|c|} \hline
Frequency&Noun&Prep&Noun2\\ \hline \hline
1983&num&to&num\\ \hline
923&num&from&num\\ \hline
853&share&from&million\\ \hline
723&trading&on&exchange\\ \hline
721&num&in&num\\ \hline
560&num&to&month\\ \hline
519&share&on&revenue\\ \hline
461&num&to&day\\ \hline
417&trading&on&yesterday\\ \hline
376&share&on&sale\\ \hline
\end{tabular}
\caption{Most frequent $(n,p,n2)$  tuples}
\label{ntuples}
\end{center}
\end{table}

%
%

\section{Statistical Models}
\label{models}

While the extracted tuples of the form $(n,p,n2)$ and $(v,p,n2)$ 
represent unambiguous noun and verb attachments in which {\em either} 
the verb or noun is known, 
our eventual goal is to resolve ambiguous attachments in the test data
of the form $(v,n,p,n2)$, in which {\em both} the noun $n$ and
verb $v$ are always known.
We therefore must use any information in the unambiguous cases
to resolve the ambiguous cases. A natural way is to
use a classifier that compares the probability of each outcome:
\begin{equation}
\begin{array}{l}
cl(v,n,p,n2) =\\
\left\{ 
\begin{array}{ll}
N&\mbox{if $p = of$}\\
\arg\max_{a \in \{N,V\}} Pr(v,n,p,a)&\mbox{otherwise}\\
\end{array}
\right.
\end{array}
\label{classifier}
\end{equation}
We do not currently use $n2$ in the probability model,
and we omit it from further discussion.

We can factor $Pr(v,n,p,a)$ as follows:
\begin{eqnarray*}
Pr(v,n,p,a) &=& Pr(v) Pr(n)\\
&&Pr(a|v,n)\\
&&Pr(p | a,v,n)
\end{eqnarray*}
The terms $Pr(n)$ and $Pr(v)$ are independent
of the attachment $a$ and need not be computed
in $cl$ (\ref{classifier}),
but the estimation of $Pr(a|v,n)$ and $Pr(p|a,v,n)$ is problematic 
since our training data, i.e., the head words extracted from
raw text, occur with either $n$ or $v$, but never both $n,v$.
This leads to make some heuristically motivated approximations.
Let the random variable $\phi$ range over $\{ true, false \}$,
and let it denote the presence or absence of any preposition
that is unambiguously attached to the noun or verb in question.
Then $p(\phi=true|n)$ is the conditional probability that
a particular noun $n$ in free text has an unambiguous prepositional phrase
attachment. ($\phi=true$ will be written simply as $true$.)
We approximate $Pr(a|v,n)$ as follows:
\begin{eqnarray*}
Pr(a=N|v,n) &\approx& {Pr(true|n) \over Z(v,n)}\\
Pr(a=V|v,n) &\approx& {Pr(true|v) \over Z(v,n)}\\
Z(v,n)      &=& Pr(true|n) + Pr(true|v)
\end{eqnarray*}
The rationale behind this approximation is that the
tendency of a $v,n$ pair towards a noun (verb) attachment
is related to the tendency of the noun (verb) alone to
occur with an unambiguous prepositional phrase.
The $Z(v,n)$ term exists only to make the approximation
a well formed probability over $a \in \{ N,V \}$.

We approximate $Pr(p|a,v,n)$ as follows:
\begin{eqnarray*}
Pr(p|a=N,v,n) &\approx& Pr(p | true, n) \\
Pr(p|a=V,v,n) &\approx& Pr(p | true, v) 
\end{eqnarray*}
The rationale behind these approximations is that 
when generating $p$ given a noun (verb) attachment, 
only the counts involving the noun (verb) are relevant,
assuming also that the noun (verb) has an attached
prepositional phrase, i.e., $\phi=true$.

We use word statistics from both the tagged corpus and the
set of extracted head word tuples to estimate the probability
of generating $\phi=true$, $p$, and $n2$.
Counts from the extracted set of tuples assume that
$\phi=true$, while counts from the corpus itself
may correspond to either $\phi=true$ or $\phi=false$,
depending on if the noun or verb in question is, or is not, respectively, 
unambiguously attached to a preposition.

\subsection{Generate $\phi$}
\label{phisection}

The quantities $Pr(true|n)$ and $Pr(true|v)$ denote the
conditional probability that $n$ or $v$ will occur with 
some unambiguously attached 
preposition,
and are estimated as follows:
\begin{eqnarray*}
Pr(true|n) &=& 
\left\{ 
\begin{array}{ll}
{c(n, true) \over c(n)}&c(n)>0\\
.5&\mbox{otherwise}\\
\end{array}
\right.\\
Pr(true|v) &=& 
\left\{ 
\begin{array}{ll}
{c(v, true) \over c(v)}&c(v)>0\\
.5&\mbox{otherwise}\\
\end{array}
\right.\\
\end{eqnarray*}
where $c(n)$ and $c(v)$ are counts
from the tagged corpus, and where
$c(n, true)$ and $c(v, true)$ are counts
from the extracted head word tuples.

\subsection{Generate $p$}

The terms  $Pr(p|n,true)$ and $Pr(p|v,true)$ denote
the conditional probability that a particular
preposition $p$ will occur as an unambiguous
attachment to $n$ or $v$.
We present two techniques to estimate 
this probability, one based on bigram counts and another
based on an interpolation method.

\subsubsection{Bigram Counts}
\label{bigramsection}

This technique uses the bigram counts of the extracted
head word tuples, and backs off to the uniform distribution
when the denominator is zero.
\begin{eqnarray*}
Pr(p|true, n) &=& 
\left\{ 
\begin{array}{ll}
{c(n, p, true) \over c(n, true)}&c(n, true)>0\\
{1 \over |\mathcal{P}|}&\mbox{otherwise}\\
\end{array}
\right.\\
Pr(p|true, v) &=& 
\left\{ 
\begin{array}{ll}
{c(v, p, true) \over c(v, true)}&c(v, true)>0\\
{1 \over |\mathcal{P}|}&\mbox{otherwise}\\
\end{array}
\right.\\
\end{eqnarray*}
where $\mathcal{P}$ is the set of possible prepositions,
where all the counts $c(\dots)$ are from the extracted head word tuples.

\subsubsection{Interpolation}
\label{interpsection}

This technique is similar to the one in \cite{hindlerooth:cl}, and  interpolates
between the tendencies of the $(v,p)$ and $(n,p)$ bigrams and the tendency of
the type of attachment (e.g., N or V) towards a particular preposition $p$.
First, define $c_N(p) =\sum_n c(n, p, true)$ 
as the number of noun attached tuples 
with the preposition $p$,
and define $c_N =  \sum_{p} c_N(p)$ 
as the number of noun attached tuples.
Analogously, define $c_V(p) =  \sum_v c(v, p, true)$ 
and  $c_V = \sum_{p} c_V(p)$.
The counts $c(n,p,true)$ and $c(v,p,true)$ are from
the extracted head word  tuples.
Using the above notation, we can interpolate as follows:
\begin{eqnarray*}
Pr(p| true, n) &=& {{c(n, p, true) + {c_N(p) \over c_N}} \over c(n, true) + 1}\\
Pr(p| true, v) &=& {{c(v, p, true) + {c_V(p) \over c_V}} \over c(v, true) + 1}\\
\end{eqnarray*}

\section{Evaluation in English}

\label{evalsec}

\newcommand\interp{cl_{interp}}
\newcommand\bigram{cl_{bigram}}
\newcommand\base{cl_{base}}

\begin{table*}
\begin{center}
\begin{tabular}{|l|l|l|l|l|} \hline
Subset&Number of Events&$\bigram$&$\interp$&$\base$\\ \hline
$p = of$&925&917&917&917\\ \hline
$p \neq of$&2172&1620&1618&1263\\ \hline
{\bf Total}&3097&2537&2535&2180\\ \hline
{\bf Accuracy}&-&81.91\%&81.85\%&70.39\%\\ \hline
\end{tabular}
\end{center}
\caption{Accuracy of mostly unsupervised classifiers on English Wall
St. Journal data}
\label{accuracy}
\end{table*}

Approximately 970K unannotated sentences from the 1988 Wall St. Journal 
were processed in a manner identical to the example
sentence in Table~\ref{table:order}.
The result was approximately 910,000 head word tuples 
of the form $(v,p,n2)$ or $(n,p,n2)$.
Note that while the head word tuples represent correct attachments
only 69\% of the time,  their quantity is about 45 times greater than 
the quantity of data used in previous supervised approaches.
The extracted data was used as training material for 
the three classifiers $\base$, $\interp$, and $\bigram$.
Each classifier is constructed as follows:
\begin{description}
\item[$\base$] This is the ``baseline'' classifier 
that predicts $N$ of $p=of$, and $V$ otherwise.

\item[$\interp$:] This classifier has the form of equation~(\ref{classifier}),
uses the method in section~\ref{phisection} to generate $\phi$, and
the method in section~\ref{interpsection} to generate $p$.

\item[$\bigram$:] This classifier has the form of equation~(\ref{classifier}),
uses the method in section~\ref{phisection} to generate $\phi$, and
the method in section~\ref{bigramsection} to generate $p$.

\end{description}
Table~\ref{accuracy} shows accuracies of the classifiers 
on the test set of \cite{maxent:ppattach}, which is derived from the 
manually annotated attachments in the Penn Treebank Wall St. Journal data.
The Penn Treebank is drawn from the 1989 Wall St. Journal data, so there is
no possibility of overlap with our training data.
Furthermore, the extraction heuristic was developed and tuned 
on a ``development set'', i.e.,  
a set of annotated examples that did not overlap with either
the test set or the training set.

\begin{table}
\begin{tabular}{|l|ll|} \hline
Attachment&$Pr(a|v,n)$&$Pr(p|a,v,n)$\\ \hline
Noun($a=N$)&.02&.24\\
Verb($a=V$)&.30&.44\\ \hline
\end{tabular}
\caption{The key probabilities for the ambiguous example {\em rise num to num}}
\label{risetable}
\end{table}

Table~\ref{risetable} shows the two
probabilities $Pr(a|v,n)$ and $Pr(p|a,v,n)$,
using the same approximations as $\bigram$,
for the ambiguous example {\em rise num to num}.
(Recall that $Pr(v)$ and $Pr(n)$ are not needed.)
While the tuple (num, to, num) is more frequent than
(rise, to, num), the conditional probabilities 
prefer $a=V$, which is the choice
that maximizes $Pr(v,n,p,a)$.

Both classifiers $\interp$ and $\bigram$ clearly outperform
the baseline, but the classifier $\interp$ does not outperform $\bigram$,
even though it interpolates 
between the less specific evidence (the preposition counts) and 
more specific evidence (the bigram counts).
This may be due to the errors in our extracted training data;
supervised classifiers that train from clean
data typically benefit greatly by combining less specific 
evidence with more specific evidence.

Despite the errors in the training data, 
the performance of the unsupervised classifiers (81.9\%) begins to approach
the best performance of the comparable supervised classifiers (84.5\%).
(Our goal is to replicate the supervision
of a treebank, but not a semantic dictionary, so
we do not compare against \cite{stetina:pp}.)
Furthermore, we do not use the second noun $n2$, 
whereas the best supervised methods use this information.
Our result shows that the information in 
imperfect but abundant data from unambiguous attachments, 
as shown in Tables~\ref{vtuples} and \ref{ntuples}, 
is sufficient to resolve ambiguous prepositional phrase
attachments at accuracies just under the supervised state-of-the-art accuracy.

\section{Evaluation in Spanish}

\begin{table*}
\begin{center}
\begin{tabular}{|l|l|l|l|l|l|} \hline
Test Set&Subset&Number of Events&$\bigram$&$\interp$&$\base$\\ \hline
All $p$&$p = de\|del$&156&154&154&154\\ \hline
&$p \neq de\|del$&116&103&97&91\\ \hline
&{\bf Total}&272&257&251&245\\ \hline
&{\bf Accuracy}&-&94.5\%&92.3\%&90.1\%\\ \hline \hline
$p=con$&{\bf Total}&192&166&160&151\\ \hline
&{\bf Accuracy}&-&86.4\%&83.3\%&78.6\%\\ \hline
\end{tabular}
\end{center}
\caption{Accuracy of mostly unsupervised classifiers on Spanish News Data}
\label{spanishaccuracy}
\end{table*}

We claim that our approach is portable to languages
with similar word order, and we support this claim 
by demonstrating our approach on the Spanish language.
We used the Spanish tagger and morphological analyzer
developed at the Xerox Research Centre Europe\footnote{These
were supplied by Dr. Lauri Kartunnen during his visit to Penn.}
and we modified the extraction heuristic to 
account for the new tagset, and to account for
the Spanish equivalents of the words {\em of} (i.e., {\em de} or {\em del})
and {\em to be} (i.e., {\em ser}). 
Chunking was not performed on the Spanish data.
We used  450k sentences of raw text from the 
Linguistic Data Consortium's Spanish News Text Collection
to extract a training set, and we used a non-overlapping set of 50k
sentences from the collection to create test sets.
Three native Spanish speakers
were asked to extract and annotate ambiguous instances
of Spanish prepositional phrase attachments.
They annotated two sets (using the full sentence context);
one  set consisted of all ambiguous prepositional phrase
attachments of the form $(v,n,p,n2)$, and the
other set consisted of cases where $p=con$.
For testing our classifier, we used only those judgments on which all three annotators agreed.

\subsection{Performance}

The performance of the classifiers $\bigram$, $\interp$, and
$\base$, when trained and tested on Spanish language data,  
are shown in Table~\ref{spanishaccuracy}. 
The Spanish test set has fewer ambiguous prepositions than the
English test set, as shown
by the accuracy of $\base$.
However, the accuracy improvements of $\bigram$ over  $\base$
are statistically significant for both test sets.\footnote{
Using proportions of changed cases, $P = 0.0258$ for 
the first set, and $P= 0.0108$ for the set where $p=con$}



%
%
%
%
\section{Conclusion}

The unsupervised algorithm for prepositional phrase attachment
presented here is the only algorithm in the published
literature that can significantly outperform
the baseline without using data derived from
a treebank or parser. 
The accuracy of our technique 
approaches the accuracy of the best supervised methods, 
and does so with only a tiny fraction of the supervision.
Since only a small part of the extraction heuristic
is specific to English, and since part-of-speech taggers
and morphology databases are widely available in other languages, our approach is 
far more portable than previous approaches for this
problem.
We successfully demonstrated the portability of our approach by
applying it to the prepositional phrase attachment task 
in the Spanish language.

\section{Acknowledgments}

We thank Dr. Lauri Kartunnen for 
lending us the Spanish natural language tools, 
and Mike Collins
for helpful discussions on this work.

\bibliography{refs}
\end{document}